\ifdefined\XeTeXversion\else\pdfoutput=1\fi
\documentclass[11pt]{article}
\usepackage[preprint]{acl}
\usepackage{times}
\usepackage[T1]{fontenc}
\usepackage[utf8]{inputenc}
\usepackage{microtype}
\usepackage{graphicx,booktabs,amsmath,amssymb,array,tabularx}
\usepackage{url}
\newcommand{\model}{Xing-Audio-NVVASR}
\newcommand{\corefine}{\textsc{Co-Refine}}
\newcommand{\tagg}[1]{\texttt{[#1]}}
\title{Xing-Audio-NVVASR: Scaling and Refining Supervision for Joint Recognition of Speech and Non-Verbal Vocalizations}
\author{Pengyu Xu, Gang Liu, Genliang Zhao, Hongjie Chen,\\
\textbf{Jian Kang, Jie Lian, Jie Li\thanks{Corresponding author.}, Yongxiang Li}\\
Xingchen AGI Lab, China Telecom Artificial Intelligence Technology Co. Ltd\\
\texttt{xupy4@chinatelecom.cn, lij86@chinatelecom.cn}}
\begin{document}
\raggedbottom
\maketitle

\begin{abstract}
Joint recognition of speech and non-verbal vocalizations (NVVs) is hindered by scarce event-rich training data and inaccurate or missing event labels. We present \model{}, an end-to-end recognizer trained through progressive data mining and iterative supervision refinement. Guided by unified acoustic criteria, two heterogeneous models cross-audit annotations, flagging disagreements for teacher reannotation. To uncover shared omissions that model agreement may conceal, complementary breath detectors screen the entire corpus for human review. Human-confirmed labels accumulate across rounds and are preserved during subsequent refinement. Over five rounds, we refine annotations for 170,000 utterances from public corpora and incorporate 630,000 mined utterances. The resulting corpus of approximately 800,000 utterances is used to train a single 1.7B-parameter model that transcribes speech with inline tags for six NVV classes. Across four human-annotated benchmarks, \model{} ranks first among nine models in both composite score and six-class Macro-F1. It achieves a mean composite score of 64.46, surpassing the strongest comparison model, StepAudio 3 Chat Preview, by 6.03 points, and exceeds the annotation teacher's breathing F1 by 14.40--31.57 percentage points. We plan to release the model weights, inference code, four evaluation benchmarks, annotation guidelines, and evaluation code.
\end{abstract}

\section{Introduction}
Non-verbal vocalizations (NVVs), such as breathing, laughter, crying, coughing, throat clearing, and sighing, complement words in natural speech \citep{xue2026nvvspeech}. NVV-aware automatic speech recognition (ASR) jointly transcribes words and vocal events, inserting event-category tags at the corresponding positions in the transcript \citep{liao2025nvspeech,ye2025nonverbalspeech,yang2026wesr}.

The NVVSpeech Challenge 2026 provides a common Mandarin--English evaluation setting for this task \citep{xue2026nvvspeech,nvvspeech2026evaluation}. However, existing public corpora underrepresent spontaneous NVVs and diverse recording conditions \citep{cao2026sourceadaptive}, while audits of automatically annotated corpora reveal systematic breathing omissions and confusion between event categories \citep{xu2026corefine}. These findings motivate expanding event coverage while improving annotation reliability. We therefore present \model{}, built through a framework that couples progressive data mining with iterative supervision refinement.

First, we expand supervision through progressive audio mining. We begin with 170,000 weakly labeled utterances from public corpora recommended by the NVVSpeech Challenge \citep{nvvspeech2026resources}: Emilia-NV/NVSpeech-170k \citep{liao2025nvspeech}, NonVerbalSpeech-38K \citep{ye2025nonverbalspeech}, and SynParaSpeech \citep{bai2026synparaspeech}. After refining this seed corpus, we progressively incorporate 630,000 utterances mined from approximately 500,000 hours of internal audio. Five rounds of refinement and expansion yield approximately 800,000 training utterances covering six NVV classes and speech without target events.

Second, we refine supervision through cross-auditing and complementary breath screening. Two heterogeneous models cross-audit the annotations, selecting disagreements for teacher reannotation and directly adopting the teacher's outputs as silver labels. If both the existing annotation and the auditing model omit an event, their transcripts can agree and the sample escapes reannotation. To detect these shared omissions and the teacher's frequent breathing misses, we screen the full corpus with Multilingual-NVASR \citep{ni2026multilingual}, whose breathing F1 exceeds the teacher's on all four benchmarks (Table~\ref{tab:breath}), and PANNs \citep{kong2020panns}, which supplies complementary utterance-level acoustic evidence. Selected candidates receive human correction under a common criterion: annotate every audibly identifiable breath without an additional intensity threshold. Human-confirmed gold labels accumulate across rounds and remain protected from automatic overwriting.

After five refinement rounds, we train a single Qwen3-ASR-1.7B recognizer \citep{shi2026qwen3asr} to generate words and six NVV tags directly from audio. Our contributions are:
\begin{itemize}
\item We develop a framework for scaling and refining six-class NVV supervision. It repairs 170,000 public-corpus utterances and progressively incorporates 630,000 mined examples through five rounds of annotation and correction, yielding approximately 800,000 training utterances.
\item We combine disagreement-based teacher annotation with full-corpus ensemble breath screening for selective human correction. The latter admits potential shared omissions into review even when cross-auditing finds agreement. Protected gold labels accumulate alongside teacher-generated silver labels across rounds.
\item We construct six-class evaluation benchmarks covering four human-annotated audio collections. We will release these benchmarks, their construction rules and annotation guidelines, and evaluation code alongside model weights and inference code.
\item Our single 1.7B-parameter recognizer ranks first in both composite score and six-class Macro-F1 among nine models on all four human-annotated benchmarks. Its mean composite score of 64.46 exceeds the strongest comparison model by 6.03 points, while breathing F1 improves over the annotation teacher by 14.40--31.57 percentage points.
\end{itemize}

\section{Related Work}
\subsection{NVV-Aware Speech Recognition}
NVV recognition has progressed from classifying isolated vocal sounds to jointly transcribing words and vocal events in natural speech. VocalSound provides data for isolated vocal-sound recognition \citep{gong2022vocalsound}. NVSpeech and NonVerbalSpeech-38K extend data construction to natural speech through scalable annotation pipelines \citep{liao2025nvspeech,ye2025nonverbalspeech}, while WESR formalizes joint word--event recognition and position-aware evaluation \citep{yang2026wesr}. The NVVSpeech Challenge brings these tasks into a bilingual setting with 16 NVV classes and a common evaluation protocol \citep{xue2026nvvspeech,nvvspeech2026resources}.

Recent progress also spans general audio-language models and task-specific recognizers. Qwen3.5-Omni-Plus supports audio understanding and reasoning within an omnimodal framework \citep{qwen2026qwen35omni}, while Gemini 3.1 Pro accepts audio alongside text, images, and video for multimodal reasoning \citep{google2026gemini31pro}. For dedicated NVV transcription, Multilingual-NVASR fine-tunes SenseVoice-Small \citep{an2024funaudiollm} to jointly transcribe Chinese or English speech and inline NVV tags under a unified paralinguistic taxonomy, and serves as a recognition component in NV-Bench evaluation \citep{ni2026nvbench,ni2026multilingual}.

As the task expands to natural speech and broader event inventories, data coverage and class imbalance become central concerns. Beyond Words studies learning for scarce NVV categories \citep{yang2026beyond}. Among challenge approaches, \citet{cao2026sourceadaptive} combine source-dependent corpus refinement with mining of spontaneous events from real-world media, while \citet{jia2026longtail} harmonize source taxonomies and combine square-root sampling with uniform-category fine-tuning. These approaches improve the availability and use of event-bearing speech. Our work addresses data coverage and annotation completeness together, coupling progressive event mining with iterative correction of erroneous and missing NVV labels.

\subsection{Learning from Noisy and Incomplete Supervision}
Learning with noisy labels addresses errors in supervision through sample selection, label-error estimation, and label correction. Co-teaching uses cross-model sample exchange to reduce memorization of noisy labels \citep{han2018coteaching}; DivideMix separates likely clean and noisy samples for semi-supervised learning \citep{li2020dividemix}; confident learning identifies likely label errors from prediction--label statistics \citep{northcutt2021confident}. Active learning complements automatic methods by selecting informative examples for human annotation \citep{settles2009active}.

In ASR, VietASR iteratively updates pretraining targets by clustering encoder features after supervised fine-tuning \citep{zhuo2025vietasr}. For NVV-aware ASR, annotation refinement must address event identity, count, and position, as well as omitted events. \corefine{} applies collaborative cross-auditing, class-aware teacher adjudication, and limited human review to iteratively refine weakly labeled NVV transcripts \citep{xu2026corefine}. However, model--label agreement can conceal shared omissions. We combine cross-auditing with full-corpus breath screening to correct erroneous labels and recover omitted events as the training corpus expands.

\section{Scaling and Refining Supervision}
\label{sec:method}
\begin{figure*}[!t]
\centering
\includegraphics[width=\textwidth]{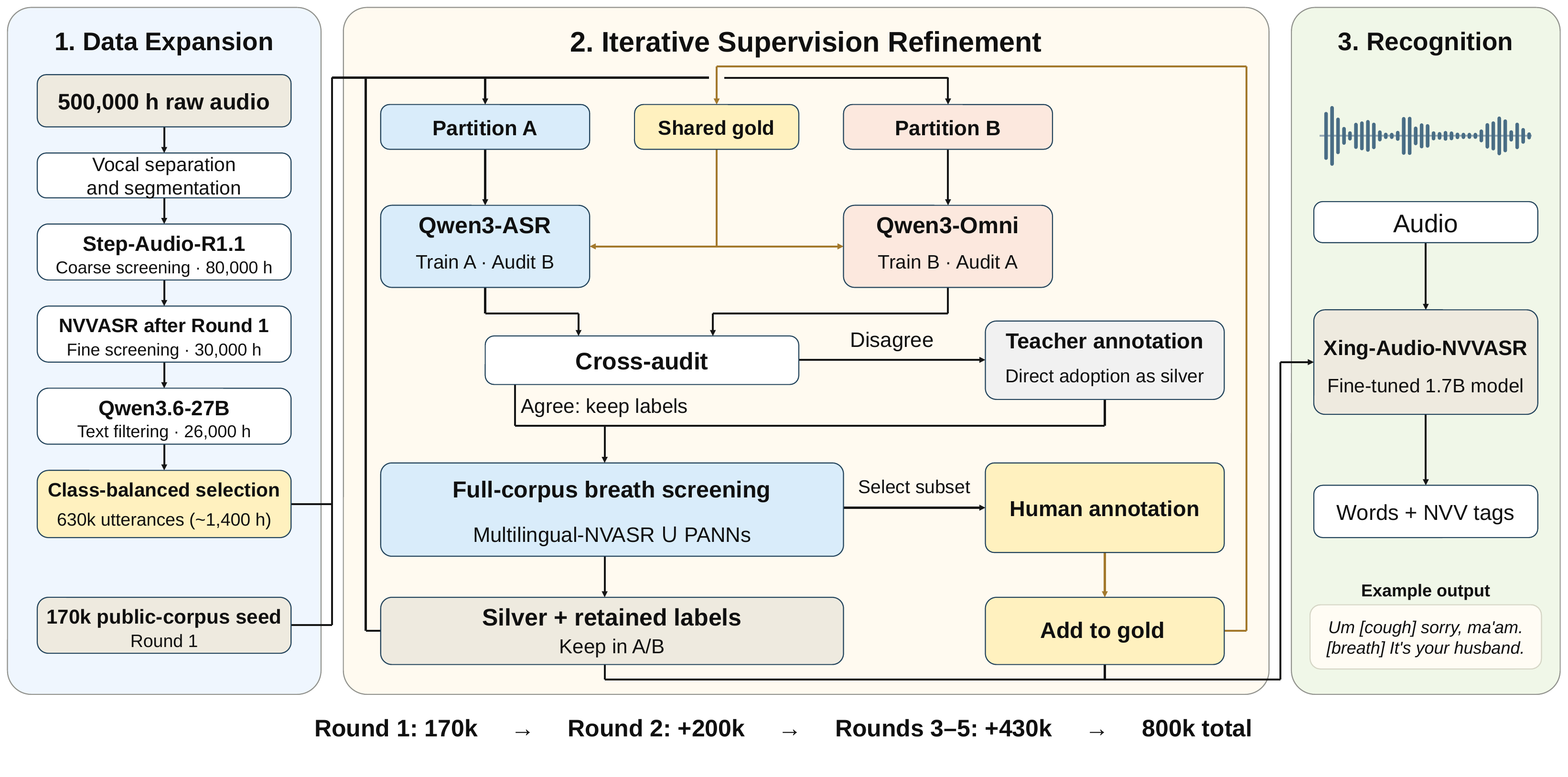}
\caption{Overview of progressive data mining and iterative supervision refinement. Cross-auditing selects disagreements for teacher reannotation, while corpus-wide ensemble breath screening identifies potential omissions for human correction. Human-confirmed gold labels remain protected across five rounds, yielding approximately 800,000 utterances for training the final recognizer.}
\label{fig:framework}
\end{figure*}

Our framework scales supervision through progressive mining and refines it through teacher reannotation and selective human correction. Two stages select samples for this correction process. Cross-auditing filters easy samples and selects disagreements for teacher annotation; ensemble breath screening covers all samples and selects potential omissions for human annotation. Teacher outputs are adopted as silver labels, and human-confirmed gold labels accumulate across rounds. We first refine 170,000 public-corpus utterances, then add 630,000 mined utterances over rounds 2--5. The resulting corpus trains the final recognizer with an event-weighted objective and distribution-guided resampling. Figure~\ref{fig:framework} summarizes the framework.

\subsection{Task and Supervision Format}
\label{sec:task}
Given audio $x$, the model generates a sequence $y$ containing lexical tokens and tags from
\begin{equation}
\begin{split}
\mathcal C=\{&\text{breath},\text{laugh},\text{cry},\text{cough},\\
&\text{throat clearing},\text{sigh}\}.
\end{split}
\end{equation}
Each occurrence of a tag denotes an event. Tags are inserted relative to the surrounding words; the output contains no start or end timestamps. For example, the illustrative output ``Well, \tagg{breath} I think we should leave'' preserves both lexical content and an event location. Repeated instances of a class remain separate events.

For analysis, write the event sequence extracted from $y$ as $E(y)=((c_j,p_j))_{j=1}^{m}$, where $c_j$ is the class and $p_j$ the number of preceding lexical units. This representation preserves event order and multiplicity.

\subsection{Data Sources and Progressive Mining}
\label{sec:mining}
\paragraph{Public-corpus seed data.}
We sample approximately 170,000 utterances from public corpora recommended by the NVVSpeech Challenge \citep{nvvspeech2026resources}, including Emilia-NV/NVSpeech-170k \citep{liao2025nvspeech}, NonVerbalSpeech-38K \citep{ye2025nonverbalspeech}, and SynParaSpeech \citep{bai2026synparaspeech}. The subset contains examples of the six target NVV classes and speech without NVV tags. Its original labels contain substantial errors \citep{xu2026corefine,cao2026sourceadaptive}, so the first round applies the full refinement procedure before these 170,000 examples support further data expansion. Speech-only examples remain in training to preserve lexical transcription and discourage unsupported event insertion.

\paragraph{Audio preprocessing.}
The mining source comprises approximately 500,000 hours of predominantly Chinese internal audio. We normalize audio formats, separate vocals using Demucs v4 \citep{rouard2023hybrid}, and construct short utterances using voice activity detection and pyannote speaker-diarization-3.1 \citep{pyannote2023diarization31}. Appendix~\ref{app:preprocessing} provides audio preparation details.

\paragraph{NVV candidate screening.}
Step-Audio-R1.1 \citep{stepfun2026stepaudior11} provides coarse paralinguistic annotation on the corresponding original-audio segments, retaining approximately 80,000 hours. The NVVASR model obtained after round 1 then performs finer event screening, retaining approximately 30,000 hours of candidate speech containing NVVs.

\paragraph{Quality filtering and class-balanced selection.}
Qwen3.6-27B \citep{qwen2026qwen3627b} evaluates the predicted transcript text, filtering incomplete sentences and anomalous transcriptions. This text-based quality filter retains approximately 26,000 hours of candidate audio. Even among the six frequent NVV classes, the candidate pool exhibits a pronounced long tail: breathing examples outnumber the other five classes combined. To address this imbalance, also studied in NVVSpeech Challenge approaches \citep{jia2026longtail}, we apply class-balanced sampling and select 630,000 utterances, totaling approximately 1,400 hours, for progressive incorporation into the refinement process.

Table~\ref{tab:construction} summarizes the mining stages.

\begin{table}[t]
\centering\small\setlength{\tabcolsep}{4pt}
\begin{tabular}{lr}
\toprule
Mining stage & Audio (h)\\
\midrule
Internal raw-audio pool & 500,000\\
After Step-Audio-R1.1 screening & 80,000\\
After NVVASR screening & 30,000\\
After text-quality filtering & 26,000\\
After class-balanced sampling & 1,400\\
\bottomrule
\end{tabular}
\caption{Approximate audio duration at each mining stage.}
\label{tab:construction}
\end{table}

Screening labels identify candidates and guide selection; complete training targets are established through the subsequent annotation and refinement rounds. Segmentation and filtering keep final training clips within 30 seconds.

\subsection{Unified Acoustic Annotation Criteria}
\label{sec:policy}
We define unified acoustic annotation criteria as a common specification of which audible events to label and how to assign them to the six target classes. These definitions are shared across data sources, annotation stages, and refinement rounds.

\paragraph{Harmonizing the initial labels.}
The 170,000 seed utterances come from public corpora whose label names and class definitions can differ. We normalize their relevant annotations to our six-class inventory: breathing, laughter, crying, coughing, throat clearing, and sighing. The first refinement round corrects labels according to these common definitions. Appendix~\ref{app:distribution} summarizes the class boundaries.

\paragraph{Maintaining definitions throughout the pipeline.}
The same class definitions guide new-data annotation, teacher annotation, and human correction. The teacher prompt remains fixed across refinement rounds. Human annotators follow the same definitions when reviewing model outputs and constructing evaluation references.

\paragraph{Labeling every audible breath.}
Our review of the selected WESR recordings reveals inconsistent inclusion of quiet breaths, reflecting differences in annotators' judgments about whether a breath is sufficiently salient (Section~\ref{sec:wesr-relabeled}). The NVVSpeech Challenge also reports persistent difficulty in recognizing breathing \citep{xue2026nvvspeech}. We therefore adopt a sensitive inclusion rule: \textbf{every breath that an annotator can hear and identify should be labeled}. This includes quiet or brief inhalation, exhalation, and ordinary speech-related respiration, without an additional intensity or duration threshold. Wind, microphone noise, and speech articulation alone do not qualify as breathing.

\subsection{Cross-Auditing and Teacher Annotation}
\label{sec:teacher}
Cross-auditing filters relatively easy examples so that the expensive teacher is used only for samples with annotation disagreements. We train Qwen3-ASR and Qwen3-Omni \citep{shi2026qwen3asr,xu2025qwen3omni} on disjoint partitions and have each model audit the opposite partition.

At round $t$, let $\mathcal D_A^{(t)}$ and $\mathcal D_B^{(t)}$ contain automatically labeled samples, and $\mathcal G^{(t)}$ contain shared, human-confirmed training samples. The two branches are
\begin{equation}
\begin{aligned}
M_A^{(t)}&=\operatorname{Train}(\text{Qwen3-ASR},
\mathcal D_A^{(t)}\cup\mathcal G^{(t)}),\\
M_B^{(t)}&=\operatorname{Train}(\text{Qwen3-Omni},
\mathcal D_B^{(t)}\cup\mathcal G^{(t)}).
\end{aligned}
\label{eq:crossfit}
\end{equation}
The initial 170,000 utterances are partitioned before round 1. Each new batch is also divided between the branches: model A provides initial labels for one half and model B for the other. They then exchange these samples for auditing before training on them. Automatically relabeled samples retain their partition membership.

For automatically labeled samples whose auditing transcripts disagree with their current annotations, Qwen3.5-Omni-Plus \citep{qwen2026qwen35omni} provides new annotations under the fixed six-class prompt. These teacher outputs directly replace the existing annotations as silver labels. Samples with matching auditing transcripts retain their current labels.

The teacher's main breathing error is omission: on XMLY-600 and XYZ-1800, breathing precision exceeds 83\%, while recall is approximately 9--11\% (Table~\ref{tab:breath}). We therefore follow teacher annotation with a separate, corpus-wide screening stage to discover missing breaths.

\subsection{Ensemble Breath Screening and Human Annotation}
\label{sec:breath-mining}
\label{sec:human}
We screen all utterances for potential breathing omissions, including both cross-audit agreements and teacher-reannotated samples, using Multilingual-NVASR and PANNs CNN14.

Multilingual-NVASR is a dedicated NVV-aware recognizer fine-tuned from SenseVoice-Small \citep{an2024funaudiollm} on six public corpora: Emilia-NV \citep{liao2025nvspeech}, NVTTS \citep{borisov2025nonverbaltts}, DisfluencySpeech \citep{wang2024disfluencyspeech}, NVS \citep{ye2025nonverbalspeech}, SMIIP-NV \citep{wu2025smiipnv}, and MNV-17 \citep{mai2025mnv17}, under a unified label taxonomy \citep{ni2026nvbench,ni2026multilingual}. It identifies breaths within the transcript and exceeds the annotation teacher, Qwen3.5-Omni-Plus, by 3.49--12.36 percentage points in breathing F1 across all four benchmarks (Table~\ref{tab:breath}).

PANNs CNN14 is pretrained for large-scale audio-event recognition \citep{kong2020panns}. In our pipeline, it only determines whether an utterance contains breathing, without transcribing words or locating events within text. Our preliminary checks show high utterance-level breathing F1, supporting its use as a sensitive candidate detector. The two models thus provide complementary evidence: NVASR identifies breaths in lexical context, while PANNs detects their acoustic presence independently of transcription.

If the transcript retained or updated after cross-auditing and teacher reannotation contains no breath tag but either detector predicts breathing, we flag the utterance for possible human review. This admits potential shared omissions even when cross-auditing finds agreement. We select a subset of these candidates for human annotation \citep{settles2009active}; annotators follow the common acoustic criteria to reject false alarms and correct event classes, counts, and transcript-relative positions.

\subsection{Iterative Refinement and Expansion}
\label{sec:rounds}
\paragraph{Gold--silver updates.}
Let $\mathcal H^{(t)}$ contain the newly human-confirmed audio--transcript pairs in round $t$. These replace their automatic counterparts and enter the shared gold set:
\begin{equation}
\mathcal G^{(t+1)}=\mathcal G^{(t)}\cup\mathcal H^{(t)}.
\label{eq:gold-update}
\end{equation}
Gold samples are removed from the automatic partitions and protected from automatic overwriting. Teacher-generated silver labels and retained automatic labels stay in their respective partitions. Each branch is retrained on its updated partition and accumulated gold samples using Eq.~\eqref{eq:crossfit}. Across five rounds, approximately 30,000 human reviews contribute to this process.

\paragraph{Progressive expansion.}
Round 1 refines the 170,000 public-corpus utterances and trains the two branches for expansion. Round 2 adds 200,000 mined utterances, with each branch initially annotating 100,000, bringing the corpus to approximately 370,000 utterances. Rounds 3--5 add the remaining 430,000 mined utterances. Each round repeats cross-auditing, teacher annotation of disagreements, full-corpus breath screening, selective human annotation, and retraining. The final corpus contains approximately 800,000 utterances: 170,000 from public corpora and 630,000 from mining, with gold, silver, and retained automatic labels.

\subsection{Training Objective and Resampling}
\label{sec:training}
After five refinement rounds, we fine-tune the final Qwen3-ASR-1.7B recognizer on the accumulated 800,000 utterances.

\paragraph{Training objective.}
Let $y_i$ be the final training transcript for audio $x_i$, with token $y_{ij}$ and prefix $y_{i,<j}$. For supervised token positions $\mathcal V$ in a batch, the objective is
\begin{equation}
\mathcal L(\theta)=
-\frac{\sum_{(i,j)\in\mathcal V} w_{ij}
\log p_\theta(y_{ij}\mid x_i,y_{i,<j})}
{\sum_{(i,j)\in\mathcal V} w_{ij}}.
\end{equation}
Because NVV tags are sparse within transcripts, equal token weighting can make lexical prediction dominate training. We therefore set $w_{ij}=2$ for target NVV tokens and $1$ for other supervised tokens to strengthen event supervision. Prompt and padding positions are masked. The objective jointly trains word and event prediction.

\paragraph{Distribution-guided resampling.}
Inspired by long-tail rebalancing in \corefine{} \citep{xu2026corefine} and NVVSpeech Challenge approaches \citep{jia2026longtail}, we set fixed sampling quotas guided by the original class distribution. Breathing and laughter receive larger quotas, while the remaining classes receive a common allocation. This retains the broad head--tail structure while narrowing the exposure gap and ensuring regular supervision for every class. Speech-only examples remain in training to maintain lexical transcription.

Sampling quotas and implementation, model configuration, optimization, validation, and inference details are provided in Appendix~\ref{app:training}.

\section{Six-Class NVV Benchmarks}
\label{sec:benchmarks}
We construct five benchmark versions covering four audio collections: WESR-400, WESR-400-Relabeled, NVVSpeech-500, XMLY-600, and XYZ-1800. Together they provide 3,306 records and approximately 8.47 hours of audio for six-class NVV-aware ASR. All references follow the NVVSpeech tagged-transcript format \citep{xue2026nvvspeech}, with event tags inserted into lexical transcripts. Table~\ref{tab:data} summarizes their scale and event distribution; Appendix~\ref{app:benchmarks} provides construction details and additional statistics.

\begin{table*}[t]
\centering\small\setlength{\tabcolsep}{3pt}
\begin{tabular}{@{}lrrrrrrrrr@{}}
\toprule
Benchmark & Records & Hours & Events & Breath & Laugh & Cry & Cough & \shortstack{Throat\\clearing} & Sigh\\
\midrule
WESR-400 & 427 & 1.037 & 583 & 172 & 224 & 27 & 99 & 29 & 32\\
WESR-400-Relabeled & 427 & 1.037 & 1,005 & 510 & 255 & 34 & 154 & 32 & 20\\
\midrule
NVVSpeech-500 & 542 & 1.582 & 846 & 170 & 71 & 211 & 170 & 167 & 57\\
XMLY-600 & 569 & 1.424 & 1,135 & 647 & 160 & 53 & 88 & 143 & 44\\
XYZ-1800 & 1,768 & 4.425 & 2,997 & 1,575 & 551 & 36 & 267 & 303 & 265\\
\bottomrule
\end{tabular}
\caption{Benchmark scale and event counts by class. Counts include repeated events within each recording. The two WESR versions share the same audio.}
\label{tab:data}
\label{tab:class-counts}
\end{table*}

\subsection{Benchmark Construction}
\label{sec:benchmark-design}
\label{sec:wesr-original-data}
\paragraph{WESR-400.}
We adapt WESR-Bench \citep{yang2026wesr} by mapping its discrete vocal-event labels to six classes and selecting compatible recordings. The resulting 427-record benchmark preserves the original event annotations. Detailed mappings and selection rules appear in Appendix~\ref{app:benchmarks}.

\paragraph{WESR-400-Relabeled.}
WESR-400-Relabeled uses the same 427 recordings as WESR-400. Multiple annotators review lexical content and vocal events under the criteria in Section~\ref{sec:policy}; their decisions are combined by voting and final correction. The audible-breath rule recovers quiet respiration omitted from the original references.

\paragraph{NVVSpeech-500.}
We annotate all 1,946 utterances in the NVVSpeech Challenge collection \citep{xue2026nvvspeech}, then retain the 542 recordings containing events from the six core classes. Crowd annotations are consolidated into the final references. This bilingual set has a relatively balanced event distribution.

\paragraph{XMLY-600.}
We construct human-annotated references for 569 Ximalaya\footnote{\url{https://www.ximalaya.com/}} recordings, extending evaluation to natural speech. Breathing accounts for 57.0\% of the target events.

\paragraph{XYZ-1800.}
We construct human-annotated references for 1,768 Xiaoyuzhou\footnote{\url{https://www.xiaoyuzhoufm.com/}} podcast recordings, extending evaluation to podcast speech. Breathing accounts for 52.6\% of the target events.

\subsection{Paired WESR References}
\label{sec:wesr-relabeled}
As shown in Table~\ref{tab:class-counts}, the reference inventory grows from 583 to 1,005 events, with breathing increasing from 172 to 510. Coughing also increases, while sighing decreases. Preserving both versions enables evaluation under the original annotations and analysis of reference completeness on identical audio. Section~\ref{sec:wesr} examines the effects on model rankings and the measured value of human supervision.

\subsection{Event Coverage}
\label{sec:benchmark-coverage}
Table~\ref{tab:class-counts} highlights three patterns: WESR relabeling mainly recovers breathing events; NVVSpeech-500 provides more balanced coverage across classes; and breathing dominates XMLY-600 and XYZ-1800. Rare classes remain sparsely represented, with only 36 crying events in XYZ-1800 despite its larger size.

\section{Experiments}
\label{sec:experiments}
\subsection{Experimental Setup}
\label{sec:eval}
We evaluate nine models on the five benchmark versions introduced in Section~\ref{sec:benchmarks}. The primary mean uses WESR-400-Relabeled, NVVSpeech-500, XMLY-600, and XYZ-1800; results on WESR-400 are reported separately for comparison with the original references.

\paragraph{Scoring protocol.}
\label{sec:metrics}
We use the six-class projection of the frozen NVVSpeech Track 1 evaluation implementation \citep{xue2026nvvspeech,nvvspeech2026evaluation}. Predicted lexical text is aligned to the reference by deterministic minimum-edit alignment, and predicted tags are mapped to reference-text coordinates. An event true positive requires a matching class and a position within a tolerance of five Chinese characters or two English words. One-to-one matching maximizes the number of matches and then minimizes position error.

For scoring group $\ell$, the composite score is
\begin{equation}
\begin{split}
S_\ell=100\big[&0.70F_{1,\ell}^{\mathrm{micro}}
+0.20(1-\mathrm{mNTD}_\ell)\\
&+0.10(1-\min(\mathrm{Err}_\ell,1))\big].
\end{split}
\label{eq:score}
\end{equation}
The mean normalized tag distance (mNTD) uses monotone event alignment and penalizes wrong classes, omissions, extra events, order, and placement. It is averaged over reference utterances with target events. Tagged-transcript error is corpus-level WER for English-mode scoring and CER for Chinese-mode scoring, with tags counted as atomic tokens.

WESR-400-Relabeled, original WESR-400, NVVSpeech-500, and XYZ-1800 average the Chinese- and English-mode scores with equal weight. XMLY-600 uses a single Chinese-mode group. Groups are assigned from reference text using the rule in Appendix~\ref{app:protocol}.

We additionally report pooled event Micro-F1 and six-class Macro-F1. The latter averages class F1 after pooling true positives, false positives, and false negatives across groups. Pooled mNTD and tagged error use their respective denominators; composite scores are computed within each language-mode group before averaging. Across benchmarks, the headline mean assigns equal weight to each of the four primary benchmark scores. Original WESR-400 is excluded from that mean. Appendix~\ref{app:protocol} details scoring groups and metric aggregation.

\paragraph{Comparison models.}
We compare with four released-weight models: Multilingual-NVASR \citep{ni2026multilingual}, MOSS-Audio-8B-Instruct \citep{yang2026moss}, Qwen3-Omni-30B-A3B-Instruct \citep{xu2025qwen3omni}, and Step-Audio-R1.1 \citep{stepfun2026stepaudior11}. The four API models are Qwen3.5-Omni-Plus \citep{qwen2026qwen35omni}, StepAudio 3 Chat Preview \citep{stepfun2026stepaudio3chat}, Doubao-Seed-2.1-Lite \citep{volcengine2026doubao21lite}, and Gemini 3.1 Pro Preview \citep{google2026gemini31pro}. General audio-language models are prompted to transcribe speech and emit the six target tags from acoustic evidence, preserving event count and order. Task-specific recognizers use their native transcription interface. The scoring pipeline projects references and predictions to the same six target classes.

In model-comparison tables, bold marks the best value and underlining marks the second-best distinct value among the displayed models. Higher is better for scores, F1, precision, and recall; lower is better for mNTD and tagged error. Ties at the reported precision receive the same marking.

\begin{table*}[t]
\centering\small\setlength{\tabcolsep}{4.5pt}\renewcommand{\arraystretch}{1.08}
\begin{tabular*}{\textwidth}{@{\extracolsep{\fill}}lrrrrrrrrr@{}}
\toprule
& \multicolumn{2}{c}{WESR-R} & \multicolumn{2}{c}{NVVSpeech} & \multicolumn{2}{c}{XMLY} & \multicolumn{2}{c}{XYZ} & Mean\\
\cmidrule(lr){2-3}\cmidrule(lr){4-5}\cmidrule(lr){6-7}\cmidrule(lr){8-9}\cmidrule(l){10-10}
Model & Score & Macro & Score & Macro & Score & Macro & Score & Macro & Score\\
\midrule
% BEGIN GENERATED table_main
Multilingual-NVASR & 51.68 & 37.44 & 38.60 & 35.32 & 39.03 & 38.57 & 45.59 & 42.75 & 43.73 \\
MOSS-Audio-8B-Instruct & 34.90 & 31.98 & 37.40 & 31.52 & 26.72 & 29.33 & 30.00 & 28.72 & 32.26 \\
Qwen3-Omni-30B-A3B & 22.94 & 17.84 & 13.61 & 19.82 & 31.23 & 33.86 & 29.91 & 29.94 & 24.42 \\
Step-Audio-R1.1 & 22.05 & 20.43 & 21.13 & 23.89 & 26.39 & 31.03 & 25.03 & 32.25 & 23.65 \\
\midrule
Qwen3.5-Omni-Plus & 55.50 & 57.47 & \underline{65.08} & \underline{63.43} & 47.45 & 55.33 & \underline{62.26} & \underline{64.86} & 57.57 \\
StepAudio 3 Chat Preview & 60.38 & 58.15 & 61.41 & 61.04 & 53.92 & \underline{61.59} & 58.00 & 64.52 & \underline{58.43} \\
Doubao-Seed-2.1-Lite & 53.11 & 51.15 & 60.51 & 58.40 & 35.78 & 42.55 & 37.94 & 42.46 & 46.84 \\
Gemini 3.1 Pro Preview & \underline{61.36} & \underline{60.67} & 57.05 & 60.18 & \underline{56.37} & 61.39 & 51.85 & 58.94 & 56.66 \\
\midrule
\textbf{Xing-Audio-NVVASR} & \textbf{66.40} & \textbf{63.91} & \textbf{68.58} & \textbf{69.20} & \textbf{58.15} & \textbf{65.24} & \textbf{64.69} & \textbf{71.73} & \textbf{64.46} \\
% END GENERATED table_main
\bottomrule
\end{tabular*}
\caption{Main results: composite score (0--100) and six-class Macro-F1 (\%). Mean averages the four benchmarks equally. Bold/underlined values mark best/second best. WESR-R denotes WESR-400-Relabeled.}
\label{tab:main}
\end{table*}

\subsection{Overall Recognition Performance}
\model{} leads the evaluated models across all four primary benchmarks in both composite score and six-class Macro-F1 (Table~\ref{tab:main}). Its mean composite score is 64.46, exceeding the strongest comparison model, StepAudio 3 Chat Preview, by 6.03 points. The composite-score margins over the strongest baseline on each benchmark range from 1.78 to 5.04 points.

The final 1.7B recognizer also exceeds its annotation teacher, Qwen3.5-Omni-Plus, by 6.88 points in mean score, and the strongest released-weight baseline, Multilingual-NVASR, by 20.73 points. The student thus outperforms its annotation teacher on the six-class recognition task.

\subsection{Breathing Recognition and Error Tradeoffs}
\begin{table*}[t]
\centering\small\setlength{\tabcolsep}{3.5pt}\renewcommand{\arraystretch}{1.08}
\begin{tabular*}{\textwidth}{@{\extracolsep{\fill}}lrrrrrrrrrrrr@{}}
\toprule
& \multicolumn{3}{c}{WESR-R} & \multicolumn{3}{c}{NVVSpeech} & \multicolumn{3}{c}{XMLY} & \multicolumn{3}{c}{XYZ}\\
\cmidrule(lr){2-4}\cmidrule(lr){5-7}\cmidrule(lr){8-10}\cmidrule(l){11-13}
Model & P & R & F1 & P & R & F1 & P & R & F1 & P & R & F1\\
\midrule
% BEGIN GENERATED table_breath
Multilingual-NVASR & \textbf{86.57} & 22.75 & 36.02 & \textbf{43.05} & 38.24 & \underline{40.50} & 73.68 & 15.15 & 25.13 & 73.38 & 13.67 & 23.04 \\
Qwen3.5-Omni-Plus & \underline{76.70} & 15.49 & 25.77 & \underline{39.78} & 21.76 & 28.14 & \textbf{92.42} & 9.44 & 17.13 & \textbf{83.50} & 11.07 & 19.55 \\
StepAudio 3 Chat Preview & 64.29 & 30.00 & 40.91 & 32.91 & 30.59 & 31.71 & 72.41 & 16.48 & 26.85 & 67.15 & 20.72 & 31.67 \\
Gemini 3.1 Pro Preview & 47.42 & \textbf{56.10} & \underline{51.40} & 20.68 & \textbf{57.65} & 30.43 & 41.66 & \textbf{54.40} & \textbf{47.18} & 35.50 & \textbf{55.53} & \textbf{43.31} \\
\textbf{Xing-Audio-NVVASR} & 69.06 & \underline{49.02} & \textbf{57.34} & 39.38 & \underline{52.35} & \textbf{44.95} & \underline{75.35} & \underline{25.04} & \underline{37.59} & \underline{74.73} & \underline{21.97} & \underline{33.95} \\
% END GENERATED table_breath
\bottomrule
\end{tabular*}
\caption{Breathing precision (P), recall (R), and F1 (\%) on four benchmarks. Bold/underlined values mark best/second best.}
\label{tab:breath}
\end{table*}

The teacher's breathing predictions are conservative: high precision accompanies severe under-detection (Table~\ref{tab:breath}). On XMLY-600 and XYZ-1800, precision exceeds 83\%, while recall is only 9.44\% and 11.07\%. This pattern motivates a separate search for omitted breaths after teacher labeling, including samples that pass cross-auditing.

\model{} achieves a better breathing precision--recall balance than the teacher on all four benchmarks. Recall gains range from 10.90 to 33.53 percentage points, accompanied by lower precision but F1 improvements of 14.40--31.57 percentage points. On XMLY-600, for example, recall rises from 9.44\% to 25.04\% and F1 from 17.13\% to 37.59\%. The improvement comes with a measurable precision cost, making both missed events and false detections relevant to further refinement.

Breathing remains a weakness on the platform recordings. Our model has the highest breathing F1 on WESR-400-Relabeled and NVVSpeech-500, whereas Gemini 3.1 Pro Preview performs better on XMLY-600 and XYZ-1800 through higher recall at lower precision. Our model recovers only about one quarter of reference breaths on XMLY-600 and one fifth on XYZ-1800. These results show that breathing recognition remains challenging in natural speech.

\subsection{Per-Class Recognition and Transcript Quality}
\begin{figure*}[t]
\centering
\includegraphics[width=\textwidth]{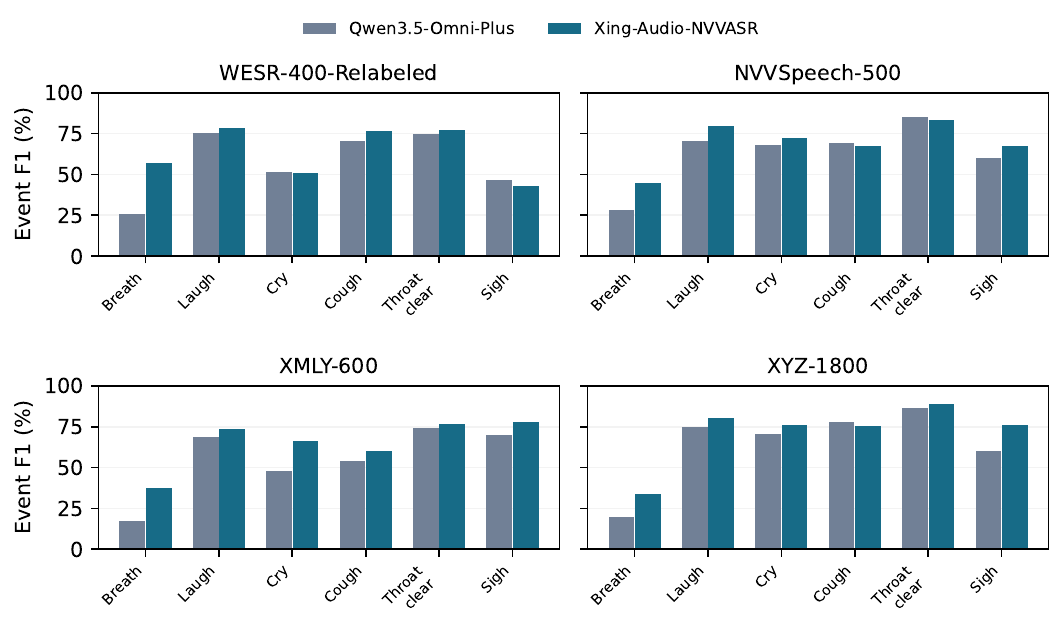}
\caption{Per-class event F1 for \model{} and the annotation teacher on all four primary benchmarks. Gains extend beyond breathing, especially to crying and sighing on the real-world sets. Coughing and throat clearing on NVVSpeech-500, and coughing on XYZ-1800, remain stronger for the teacher.}
\label{fig:classes}
\end{figure*}

The gains extend beyond breathing: our model outperforms the teacher in crying and sighing on both XMLY-600 and XYZ-1800 (Figure~\ref{fig:classes}). Crying F1 improves by 18.15 percentage points on XMLY-600, and sighing F1 by 15.36 points on XYZ-1800. These gains do not extend to every benchmark: crying and sighing remain below the teacher on WESR-400-Relabeled. Coughing on NVVSpeech-500 and XYZ-1800, and throat clearing on NVVSpeech-500, also remain weaker.

Better event recognition does not always yield more accurate full transcripts. Compared with the teacher, mNTD decreases on all four benchmarks, but tagged-transcript error decreases on only three. On XYZ-1800, tagged-transcript error rises from 5.13\% to 6.10\% despite the improvement in mNTD.

\subsection{WESR: Test-Set Omissions and Relabeling}
\label{sec:wesr}
During our experiments, adding 24,000 human-annotated training utterances unexpectedly reduced performance on WESR-400. With other training conditions fixed, the composite score fell from 72.69 to 65.10, and breathing F1 fell from 57.23\% to 47.94\% (Table~\ref{tab:motivation}). This raised a question: why did additional human supervision lead to substantially worse benchmark results?

To investigate, we listened to the test recordings and checked their reference transcripts. We found substantial annotation omissions, particularly for breathing: many audible breaths had no corresponding tag. These omissions can cause correct breathing predictions to be counted as false positives, making the original test set unreliable for assessing improvements in breathing detection. We therefore relabeled the same recordings using our audible-breath criterion, producing WESR-400-Relabeled. The number of annotated breathing events increased from 172 to 510 (Section~\ref{sec:wesr-relabeled}).

On the relabeled test set, the model trained with the additional human data performs better: its composite score is 66.40 versus 60.43 without that data, and its breathing F1 is 57.34\% versus 36.50\%. The augmented model is \model{}; full per-class results are provided in Appendix~\ref{app:human-comparison}. The initial performance decline thus led us to identify and correct substantial gaps in the test annotations. WESR-400-Relabeled provides a more complete reference for evaluating whether models recognize the audible events targeted by our annotation criterion.

\begin{table}[t]
\centering\small\setlength{\tabcolsep}{3pt}
\begin{tabular}{lrrr}
\toprule
Added human data & 0 & 24k & Change\\
\midrule
\multicolumn{4}{l}{\textit{Original WESR-400 references}}\\
Composite score & \textbf{72.69} & \underline{65.10} & \textminus 7.59 \\
Breath F1 & \textbf{57.23} & \underline{47.94} & \textminus 9.29 \\
\midrule
\multicolumn{4}{l}{\textit{WESR-400-Relabeled references}}\\
Composite score & \underline{60.43} & \textbf{66.40} & +5.97 \\
Breath F1 & \underline{36.50} & \textbf{57.34} & +20.84 \\
\bottomrule
\end{tabular}
\caption{Effect of adding 24,000 human-annotated training utterances under two WESR reference versions, with other training conditions fixed. Scores use a 0--100 scale; F1 is in percent.}
\label{tab:motivation}
\end{table}

\section{Discussion}
\subsection{Revisiting Breathing Recognition Difficulties}
\label{sec:challenge-response}
The NVVSpeech Challenge overview identifies breath and gasp among the weakest classes in its class-wise recognition analysis \citep[Section~4.1]{xue2026nvvspeech}. Our findings respond to this observation for breathing by highlighting the role of annotation quality in both evaluation and training.

Incomplete test annotations can make a model that detects more real breaths appear worse. A breath that is audible but absent from the reference may be detected correctly and still be scored as an error. We observe this problem on WESR-400: adding human-annotated training data reduces breathing F1 by 9.29 percentage points under the original test labels, but increases it by 20.84 points after the same test audio is relabeled (Table~\ref{tab:motivation}). The models are unchanged; only the reference annotations differ. This comparison shows why low breathing scores must be examined for missing reference labels before being taken as evidence of poor recognition.

Our training-data audits also reveal widespread breathing omissions, leaving models without sufficient explicit supervision for audible events. Recovering these labels is therefore central to our breath screening and human correction. Breathing recognition still needs improvement, especially in recall on natural recordings, but its reported weakness calls for closer study of annotation errors alongside acoustic modeling. In response to the challenge findings, we emphasize two priorities: reliable test references and adequate, consistently labeled breathing examples for training.

\subsection{Annotate Every Audible Breath}
In our annotation work with multiple annotators, we observed that breathing labels could differ substantially across annotators. These differences appeared more pronounced for breathing than for other vocalization classes, suggesting that subjective judgment plays a particularly important role in breathing annotation. This observation highlights the difficulty of obtaining consistent human labels for breathing and the need for clear, shared annotation criteria.

Our experience suggests two practical measures for improving breathing annotations. First, the guidelines should be clear and directly actionable. We use the same rule in training-data refinement and benchmark annotation: every event that can be heard and identified as breathing must be labeled, including faint breaths and breathing between phrases, without an additional intensity threshold. This avoids asking annotators to decide whether a breath is ``prominent enough.'' Teacher prompts and human annotation guidelines follow the same criterion.

Second, annotations should be collected from multiple annotators and consolidated into a final reference. In our WESR relabeling, we combine their decisions through voting and final correction. This reduces reliance on any one annotator's judgment and allows disagreements to be reviewed. Clear guidelines and consolidation of multiple annotations together help reduce omissions and inconsistent labeling.

\subsection{Data Quality and Selection}
\label{sec:data-value}
A recent challenge study achieves a score of 63.86 using only 26,648 training utterances \citep{jia2026longtail}, just 3.71\% below the highest score reported in the challenge overview \citep{xue2026nvvspeech}. This result suggests that a relatively small, carefully prepared training set can support competitive NVV-aware recognition.

Our development experiments show a similar pattern: approximately 50,000 high-quality training utterances already support strong recognition, whereas much larger collections with poor annotations yield substantially weaker performance. These observations suggest that increasing data volume cannot reliably compensate for incomplete or incorrect supervision.

These observations motivate future work on data valuation and selection \citep{ghorbani2019datashapley,paul2021datadiet} to identify high-quality training subsets and explore the performance limits achievable with them.

\section{Conclusion}
We presented \model{}, a 1.7B recognizer for joint speech and six-class NVV transcription. Cross-auditing selects disagreements for teacher reannotation, while full-corpus breath screening identifies potential shared omissions for human correction; confirmed gold labels remain protected across rounds. Combined with progressive mining, five rounds of refinement expand 170,000 weakly labeled utterances into approximately 800,000 training examples. Across four human-annotated benchmarks, the model ranks first among nine models in composite score and six-class Macro-F1, exceeding the strongest baseline by 6.03 points in mean score and its annotation teacher by 14.40--31.57 percentage points in breathing F1. We will release model weights, four evaluation benchmarks, and supporting code and annotation guidelines.

\section*{Limitations}
\paragraph{Remaining annotation omissions.}
Substantial event omissions remain in our training data. Targeted correction of only part of the missing annotations has already yielded substantial gains, but our refinement process does not recover all omitted events. The gains achievable through more complete annotation correction remain unexplored.

\paragraph{Unassessed data value.}
We have not evaluated the contribution of individual examples within the approximately 800,000-utterance fine-tuning set. We suspect that it contains substantial redundancy, examples of limited value, and some that may harm performance. The current training procedure uses the dataset as a whole without selection based on sample value; these possibilities and the benefits of such selection remain untested.

\paragraph{Limited event coverage.}
This initial study covers only six common categories of discrete non-verbal vocalizations. Rare, long-tail categories remain outside its scope, as do continuous vocal events mixed with speech, as distinguished from standalone discrete events in WESR \citep{yang2026wesr}. Extending annotation and recognition to these types remains future work.

\bibliography{references}

\clearpage
\appendix
\section{Data Construction and Annotation}
\subsection{Audio Preparation for NVV Mining}
\label{app:preprocessing}
For NVV mining, Demucs v4 separates vocals from background accompaniment \citep{rouard2023hybrid}. Silero VAD \citep{silero2024vad} and pyannote speaker-diarization-3.1 \citep{pyannote2023diarization31} provide speech-activity and speaker boundaries for splitting and merging recordings into short utterances. We preserve the correspondence between original and separated audio: the separated stream supports segmentation, while Step-Audio-R1.1 receives the corresponding original-audio segments for vocal-event annotation. Screening and refinement of the selected utterances follow Section~\ref{sec:mining} and Sections~\ref{sec:teacher}--\ref{sec:rounds}.

\subsection{Acoustic Annotation Guidelines}
\label{app:distribution}
Table~\ref{tab:annotation-criteria} summarizes the six-class criteria used in teacher prompts, human correction, and newly constructed benchmark annotations. Four distinctions require particular attention. For breathing, we adopt a sensitive inclusion rule: every audible, identifiable breath must be labeled, without an intensity or duration threshold. Laughter and crying should be distinguished from merely cheerful or sad speech; the corresponding tone alone does not establish a vocal event. Coughing and throat clearing are easily confused and must be distinguished from each other. Sighing is an expressive exhalation that can be confused with ordinary prolonged exhalation; our criteria therefore require an audible expressive quality rather than duration alone.

Annotators preserve event count and order and place tags relative to the spoken words. Speech-only recordings retain speech-only transcripts; recordings with target events but no words may contain tags alone.

\begin{table}[t]
\centering\small\setlength{\tabcolsep}{6pt}
\begin{tabularx}{\columnwidth}{@{}l>{\raggedright\arraybackslash}X@{}}
\toprule
Class & Acoustic criterion\\
\midrule
Breath & Audible, identifiable inhalation, exhalation, or speech-related respiration, including quiet and brief breaths. No additional intensity or duration threshold applies. Wind, microphone noise, and speech articulation alone do not qualify.\\
Laugh & Audible laughter, including chuckles, giggles, and crowd laughter. Cheerful speech alone is insufficient.\\
Cry & Audible sobbing, whimpering, wailing, or a clearly crying voice. A sad tone or sad linguistic content alone is insufficient.\\
Cough & An audible explosive cough, distinguished from throat clearing.\\
Throat clearing & Audible clearing of the throat, distinguished from coughing.\\
Sigh & An audible expressive breath release. A spoken interjection or a long exhalation alone is insufficient.\\
\bottomrule
\end{tabularx}
\caption{Six-class acoustic annotation criteria.}
\label{tab:annotation-criteria}
\end{table}

\section{Training and Inference Configuration}
\label{app:training}
We fully fine-tune Qwen3-ASR-1.7B after round 5, adding six atomic NVV tokens. Training retains speech-only examples, excludes annotations with non-target or invalid tags, and caps audio inputs at 30 seconds. The training input rate is 16 kHz; upstream preprocessing uses 24 kHz.

\begin{table}[t]
\centering\small\setlength{\tabcolsep}{6pt}
\begin{tabularx}{\columnwidth}{@{}l>{\raggedright\arraybackslash}X@{}}
\toprule
Setting & Configuration\\
\midrule
Hardware & Eight NVIDIA A800 80GB GPUs\\
Batch size & 8 per GPU; 1 gradient-accumulation step; global batch 64\\
Learning rate & 0.00002; 500 warmup steps; cosine decay\\
Regularization & Weight decay 0.01; gradient clipping 1.0\\
Precision and memory & BF16; gradient checkpointing\\
Sampling budget & Up to 50 cycles of 20,000 utterance exposures\\
Checkpoint schedule & Evaluate and save every 500 steps; retain the top three by validation composite score\\
Early stopping & Ten evaluations without improvement\\
Random seed & 42\\
\bottomrule
\end{tabularx}
\caption{Configured final-recognizer training settings. The token-weighted objective is given in Section~\ref{sec:training}.}
\label{tab:training-config}
\end{table}

\paragraph{Sampling.}
Our observations on this NVV recognition task suggest that both the original long-tailed distribution and a fully balanced distribution are suboptimal. Sampling from the original distribution severely underrepresents rare classes, while full balancing departs from the class frequencies encountered at evaluation. We therefore use a conservative rebalancing strategy that increases exposure to rare classes while retaining larger quotas for common events. The quotas are chosen empirically to moderate the imbalance while preserving part of the original distribution.

Each cycle draws 20,000 utterance exposures using the fixed quotas in Table~\ref{tab:quotas}: 12,000 from NVV-bearing pools and 8,000 from speech without target tags. Multi-class samples enter the pool of their least frequent label, with ties resolved by a fixed class order. Pools smaller than their quota are sampled with replacement. Quotas count sampled exposures rather than distinct utterances.

\begin{table}[ht]
\centering\small
\begin{tabular}{lr}
\toprule
Sampling pool & Exposures per cycle\\
\midrule
Breath & 4,000\\
Laugh & 2,000\\
Cry, cough, throat clearing, sigh & 1,500 each\\
Speech without target tags & 8,000\\
\midrule
Total & 20,000\\
\bottomrule
\end{tabular}
\caption{Final-model sampling quotas per cycle. Exposures count sampled utterances.}
\label{tab:quotas}
\end{table}

\paragraph{Validation.}
Validation is split from the refined training snapshot by available audio identity and source-recording groups. Target quotas are 100 breathing examples, 50 per remaining class, and 100 speech-only examples; group constraints can reduce the realized counts.

\paragraph{Inference.}
Evaluation uses \texttt{qwen-asr} with the Transformers backend, BF16, seed 42, greedy decoding, automatic language identification, and a 256-token generation cap. The optional vLLM bulk-labeling path has a 512-token default. Teachers and breath-screening models are used for data construction; deployed recognition uses the single fine-tuned model.

\section{Benchmark Construction and Statistics}
\label{app:benchmarks}
\subsection{WESR Mapping and Selection}
WESR covers a broader range of paralinguistic categories, including both discrete events and continuous spans, whereas this study focuses on six core NVV classes. We therefore map compatible discrete labels to these classes (Table~\ref{tab:wesr-mapping}) and select recordings that fit our event-based transcription format. The selected collection includes 34 crowd-laughter events in 28 recordings, consistent with the annotation criteria in Appendix~\ref{app:distribution}.

\begin{table}[t]
\centering\small\setlength{\tabcolsep}{6pt}
\begin{tabularx}{\columnwidth}{@{}l>{\raggedright\arraybackslash}X@{}}
\toprule
Target class & WESR discrete source labels\\
\midrule
Breath & \texttt{[inhale]}, \texttt{[exhale]}, \texttt{[breathing]}\\
Laugh & \texttt{[laughs]}, \texttt{[laughing]}, \texttt{[crowd\_laughter]}, \texttt{[chuckle]}, \texttt{[giggle]}\\
Cry & \texttt{[cry]}, \texttt{[sobbing]}\\
Cough & \texttt{[cough]}\\
Throat clearing & \texttt{[clear\_throat]}\\
Sigh & \texttt{[sigh]}\\
\bottomrule
\end{tabularx}
\caption{Mapping of discrete WESR events to the six target classes. Tags retain their transcript-relative positions.}
\label{tab:wesr-mapping}
\end{table}

\paragraph{Selection.}
Each selected record contains at least one mapped discrete event. Recordings with continuous crying, laughing, or panting spans are excluded to avoid converting sustained events into single discrete tags. Non-target style spans, such as whispering, shouting, and singing, are removed while their words are retained. Records with unsupported discrete events, unknown markers, or malformed markup are excluded.

The resulting benchmark contains 427 recordings. Retained records preserve source split, row, original discrete labels, and removed non-target span labels. Source metadata identifies 218 Chinese and 209 English records; the text-based scoring rule in Appendix~\ref{app:protocol} assigns 224 to Chinese mode and 203 to English mode. The two WESR versions retain matched audio identifiers and separate reference files.

\subsection{Human Annotation and Audio Statistics}
Because the NVVSpeech Challenge organizers provided audio without reference labels, we manually annotated all 1,946 utterances and selected the 542 recordings containing the six target classes to construct our test set. Supplied language metadata identifies 285 Chinese and 257 English records. XMLY-600 and XYZ-1800 also use manually annotated transcripts with six-class event labels.

WESR durations range from 3.02 to 29.97 seconds (median 6.83); XMLY ranges from 2.05 to 24.33 seconds (median 7.51); XYZ ranges from 1.00 to 37.06 seconds (median 5.81). Table~\ref{tab:benchmark-details} summarizes the evaluated inventories.

\begin{table*}[tp]
\centering\small\setlength{\tabcolsep}{4pt}
\begin{tabular}{lrrrrrr}
\toprule
Benchmark & Records & Audio (h) & Mean (s) & Events & Multi-class (\%) & No event\\
\midrule
WESR-400 & 427 & 1.037 & 8.75 & 583 & 10.8 & 0\\
WESR-400-Relabeled & 427 & 1.037 & 8.75 & 1,005 & 39.1 & 4\\
NVVSpeech-500 & 542 & 1.582 & 10.51 & 846 & 14.4 & 0\\
XMLY-600 & 569 & 1.424 & 9.01 & 1,135 & 27.8 & 22\\
XYZ-1800 & 1,768 & 4.425 & 9.01 & 2,997 & 23.4 & 143\\
\midrule
Four-collection total & 3,306 & $\approx$8.47 & --- & 5,983 & --- & 169\\
\bottomrule
\end{tabular}
\caption{Five benchmark versions drawn from four audio collections. The total counts the WESR audio once, using its relabeled event inventory. Dataset-name suffixes are approximate; record counts are exact for the evaluated versions. Durations are measured from audio, events count tag occurrences, and multi-class records contain at least two distinct target classes.}
\label{tab:benchmark-details}
\end{table*}

Relabeling increases the proportion of multi-class WESR records from 10.8\% to 39.1\%. Breathing contributes 80.1\% of the net increase in event count. The four primary benchmarks contain 169 records without target events, allowing false-insertion evaluation.

\section{Evaluation Protocol and Aggregation}
\label{app:protocol}
\paragraph{Implementation.}
The six-class scorer uses the official NVVSpeech Challenge Track 1 implementation \citep{xue2026nvvspeech,nvvspeech2026evaluation}. It retains the official alignment and score formula and removes recognized non-target tags.

\paragraph{Scoring groups.}
A reference containing a Chinese character is assigned to Chinese-mode scoring; all other references, including event-only transcripts, use English-mode scoring. This text rule determines tokenization and grouping independently of language metadata and the recognizer's automatic language identification. Composite scores average the available groups equally, as specified in Section~\ref{sec:metrics}.

\paragraph{Pooling.}
For group-wise event counts $\mathrm{TP}_\ell$, $\mathrm{FP}_\ell$, and $\mathrm{FN}_\ell$, pooled event F1 is
\begin{equation}
F_1^{\mathrm{pool}}=
\frac{2\sum_\ell\mathrm{TP}_\ell}
{2\sum_\ell\mathrm{TP}_\ell+\sum_\ell\mathrm{FP}_\ell+\sum_\ell\mathrm{FN}_\ell}.
\end{equation}
Class F1 pools counts separately for each class; Macro-F1 averages the six values. Pooled mNTD is weighted by the number of reference utterances with target events. Pooled tagged error divides total edit distance by total reference-token count, using each group's tokenization. Composite scores are computed within groups before averaging, rather than reconstructed from pooled metrics.

\begin{table}[t]
\centering\small\setlength{\tabcolsep}{3pt}
\begin{tabular}{@{}lrrr@{}}
\toprule
Benchmark & \shortstack{Chinese\\mode} & \shortstack{English\\mode} & Score\\
\midrule
WESR-400-Relabeled & 60.80 & 71.99 & 66.40\\
NVVSpeech-500 & 69.07 & 68.09 & 68.58\\
XMLY-600 & 58.15 & --- & 58.15\\
XYZ-1800 & 63.99 & 65.40 & 64.69\\
\midrule
WESR-400 & 65.48 & 64.73 & 65.10\\
\bottomrule
\end{tabular}
\caption{Language-mode scores for \model{} (0--100). Group scores and composites are rounded separately.}
\label{tab:score-groups}
\end{table}

\section{Supplementary Recognition Results}
All results use the protocol in Appendix~\ref{app:protocol}. Scores are on a 0--100 scale; precision, recall, and F1 are percentages. mNTD and tagged error are multiplied by 100, with lower values indicating better performance. Bold and underlining mark the best and second-best values in each column. Throat denotes throat clearing.

\subsection{Results on the Five Benchmark Versions}
\label{app:details}
Table~\ref{tab:detail-wesro} reports original WESR-400 results, followed by the four primary benchmarks in Tables~\ref{tab:detail-wesrr}--\ref{tab:detail-xyz}. Original WESR-400 shares audio with its relabeled version and is excluded from the primary mean.

\begin{table*}[tp]
\centering\small\setlength{\tabcolsep}{4.5pt}\renewcommand{\arraystretch}{1.08}
\begin{tabular*}{\textwidth}{@{\extracolsep{\fill}}lrrrrrrrrr@{}}
\toprule
Model & Micro-F1 & mNTD$\downarrow$ & Err$\downarrow$ & Breath & Laugh & Cry & Cough & Throat & Sigh\\
\midrule
% BEGIN GENERATED table_detail_WESR-400
Multilingual-NVASR & 61.69 & 45.10 & 13.02 & \textbf{55.02} & 74.67 & 40.00 & 68.00 & 0.00 & 28.57 \\
MOSS-Audio-8B-Instruct & 40.04 & 65.56 & 12.55 & 18.27 & 50.90 & 20.65 & 64.74 & 46.81 & 26.92 \\
Qwen3-Omni-30B-A3B & 15.89 & 66.80 & 25.63 & 11.02 & 47.62 & 10.95 & 8.66 & 21.90 & 11.21 \\
Step-Audio-R1.1 & 29.53 & 71.54 & 91.05 & 15.04 & 40.57 & 8.16 & 35.80 & 30.19 & 12.77 \\
\midrule
Qwen3.5-Omni-Plus & \textbf{67.17} & \underline{37.89} & \textbf{9.36} & \underline{48.42} & 78.54 & \textbf{54.90} & \textbf{75.13} & \textbf{80.70} & 52.94 \\
StepAudio 3 Chat Preview & \underline{64.85} & \textbf{35.30} & 11.34 & 47.69 & \textbf{82.92} & 45.90 & 63.41 & 68.97 & \underline{58.82} \\
Doubao-Seed-2.1-Lite & 55.87 & 50.77 & \underline{11.02} & 39.77 & 70.37 & 40.00 & 63.78 & 50.00 & 54.90 \\
Gemini 3.1 Pro Preview & 50.73 & 55.96 & 16.79 & 32.53 & 73.88 & 37.11 & 56.10 & \underline{78.57} & 58.06 \\
\midrule
\textbf{Xing-Audio-NVVASR} & 63.06 & 39.20 & 11.09 & 47.94 & \underline{81.17} & \underline{48.28} & \underline{70.00} & 74.58 & \textbf{59.74} \\
% END GENERATED table_detail_WESR-400
\bottomrule
\end{tabular*}
\caption{Original-reference WESR pooled diagnostics and per-class F1.}
\label{tab:detail-wesro}
\end{table*}

\begin{table*}[tp]
\centering\small\setlength{\tabcolsep}{4.5pt}\renewcommand{\arraystretch}{1.08}
\begin{tabular*}{\textwidth}{@{\extracolsep{\fill}}lrrrrrrrrr@{}}
\toprule
Model & Micro-F1 & mNTD$\downarrow$ & Err$\downarrow$ & Breath & Laugh & Cry & Cough & Throat & Sigh\\
\midrule
% BEGIN GENERATED table_detail_WESR-400-Relabeled
Multilingual-NVASR & 48.42 & 57.55 & 15.42 & 36.02 & 71.50 & 33.33 & 53.75 & 0.00 & 30.00 \\
MOSS-Audio-8B-Instruct & 29.33 & 73.70 & 15.00 & 8.79 & 47.12 & 23.46 & 53.51 & 44.00 & 15.00 \\
Qwen3-Omni-30B-A3B & 15.17 & 74.88 & 27.34 & 10.14 & 44.33 & 14.42 & 9.85 & 22.86 & 5.45 \\
Step-Audio-R1.1 & 22.68 & 79.09 & 90.75 & 7.62 & 40.46 & 7.14 & 27.65 & 27.52 & 12.20 \\
\midrule
Qwen3.5-Omni-Plus & 51.90 & 53.36 & \underline{11.63} & 25.77 & 75.57 & \underline{51.72} & 70.49 & \underline{74.58} & 46.67 \\
StepAudio 3 Chat Preview & 58.76 & 47.97 & 12.86 & 40.91 & \textbf{82.10} & 41.18 & 68.16 & 72.13 & 44.44 \\
Doubao-Seed-2.1-Lite & 49.20 & 56.53 & 12.74 & 36.07 & 66.01 & \textbf{53.06} & 60.18 & 40.43 & \underline{51.16} \\
Gemini 3.1 Pro Preview & \underline{60.38} & \underline{46.52} & 16.18 & \underline{51.40} & 74.58 & 38.33 & \underline{75.88} & 70.00 & \textbf{53.85} \\
\midrule
\textbf{Xing-Audio-NVVASR} & \textbf{65.15} & \textbf{43.54} & \textbf{11.28} & \textbf{57.34} & \underline{78.18} & 51.06 & \textbf{76.36} & \textbf{77.42} & 43.08 \\
% END GENERATED table_detail_WESR-400-Relabeled
\bottomrule
\end{tabular*}
\caption{WESR-400-Relabeled pooled diagnostics and per-class F1.}
\label{tab:detail-wesrr}
\end{table*}

\begin{table*}[tp]
\centering\small\setlength{\tabcolsep}{4.5pt}\renewcommand{\arraystretch}{1.08}
\begin{tabular*}{\textwidth}{@{\extracolsep{\fill}}lrrrrrrrrr@{}}
\toprule
Model & Micro-F1 & mNTD$\downarrow$ & Err$\downarrow$ & Breath & Laugh & Cry & Cough & Throat & Sigh\\
\midrule
% BEGIN GENERATED table_detail_NVVSpeech-500
Multilingual-NVASR & 35.63 & 71.32 & 21.18 & \underline{40.50} & 56.30 & 19.01 & 46.12 & 0.00 & 50.00 \\
MOSS-Audio-8B-Instruct & 34.12 & 74.64 & 19.35 & 13.33 & 32.65 & 37.39 & 38.74 & 53.04 & 13.95 \\
Qwen3-Omni-30B-A3B & 6.66 & 61.62 & 112.08 & 2.13 & 58.65 & 4.61 & 4.07 & 19.27 & 30.21 \\
Step-Audio-R1.1 & 24.25 & 80.14 & 108.79 & 8.90 & 24.49 & 2.65 & 27.82 & 47.17 & 32.31 \\
\midrule
Qwen3.5-Omni-Plus & \underline{64.05} & \underline{42.42} & \underline{15.20} & 28.14 & \underline{70.23} & 67.76 & \textbf{69.26} & \textbf{85.33} & \underline{59.85} \\
StepAudio 3 Chat Preview & 60.61 & 46.58 & 17.32 & 31.71 & 67.05 & 62.32 & 64.03 & \underline{83.23} & 57.89 \\
Doubao-Seed-2.1-Lite & 59.65 & 48.81 & 17.36 & 25.74 & 62.90 & \underline{68.68} & \underline{68.61} & 74.46 & 50.00 \\
Gemini 3.1 Pro Preview & 56.19 & 49.40 & 23.78 & 30.43 & 65.17 & 66.13 & 62.65 & 81.48 & 55.21 \\
\midrule
\textbf{Xing-Audio-NVVASR} & \textbf{67.56} & \textbf{36.51} & \textbf{13.55} & \textbf{44.95} & \textbf{79.73} & \textbf{72.57} & 67.27 & 83.13 & \textbf{67.57} \\
% END GENERATED table_detail_NVVSpeech-500
\bottomrule
\end{tabular*}
\caption{NVVSpeech-500 pooled diagnostics and per-class F1.}
\label{tab:detail-nvv}
\end{table*}

\begin{table*}[tp]
\centering\small\setlength{\tabcolsep}{4.5pt}\renewcommand{\arraystretch}{1.08}
\begin{tabular*}{\textwidth}{@{\extracolsep{\fill}}lrrrrrrrrr@{}}
\toprule
Model & Micro-F1 & mNTD$\downarrow$ & Err$\downarrow$ & Breath & Laugh & Cry & Cough & Throat & Sigh\\
\midrule
% BEGIN GENERATED table_detail_XMLY-600
Multilingual-NVASR & 34.56 & 68.76 & 14.05 & 25.13 & 66.92 & 35.82 & 35.57 & 0.00 & 67.96 \\
MOSS-Audio-8B-Instruct & 21.27 & 84.36 & 13.02 & 4.45 & 28.85 & 56.25 & 39.23 & 40.44 & 6.78 \\
Qwen3-Omni-30B-A3B & 26.11 & 76.04 & 18.38 & 5.44 & 48.58 & 47.31 & 36.51 & 46.81 & 18.52 \\
Step-Audio-R1.1 & 25.86 & 75.54 & 66.03 & 5.38 & 34.76 & 3.51 & 15.24 & 60.65 & 66.67 \\
\midrule
Qwen3.5-Omni-Plus & 43.97 & 62.02 & \underline{9.26} & 17.13 & 68.82 & 47.83 & 54.19 & 74.19 & \underline{69.81} \\
StepAudio 3 Chat Preview & 50.96 & 53.84 & 9.82 & 26.85 & \textbf{75.69} & \underline{63.27} & 56.79 & \textbf{78.26} & 68.69 \\
Doubao-Seed-2.1-Lite & 31.44 & 75.73 & 10.83 & 11.40 & 50.22 & 50.00 & 37.84 & 57.46 & 48.39 \\
Gemini 3.1 Pro Preview & \underline{54.46} & \underline{50.44} & 16.64 & \textbf{47.18} & 67.25 & 55.28 & \underline{58.23} & 71.97 & 68.42 \\
\midrule
\textbf{Xing-Audio-NVVASR} & \textbf{54.86} & \textbf{47.11} & \textbf{8.30} & \underline{37.59} & \underline{73.33} & \textbf{65.98} & \textbf{60.11} & \underline{76.54} & \textbf{77.88} \\
% END GENERATED table_detail_XMLY-600
\bottomrule
\end{tabular*}
\caption{XMLY-600 pooled diagnostics and per-class F1.}
\label{tab:detail-xmly}
\end{table*}

\begin{table*}[tp]
\centering\small\setlength{\tabcolsep}{4.5pt}\renewcommand{\arraystretch}{1.08}
\begin{tabular*}{\textwidth}{@{\extracolsep{\fill}}lrrrrrrrrr@{}}
\toprule
Model & Micro-F1 & mNTD$\downarrow$ & Err$\downarrow$ & Breath & Laugh & Cry & Cough & Throat & Sigh\\
\midrule
% BEGIN GENERATED table_detail_XYZ-1800
Multilingual-NVASR & 41.79 & 57.84 & 10.46 & 23.04 & 71.91 & 43.86 & 47.03 & 0.00 & \underline{70.67} \\
MOSS-Audio-8B-Instruct & 21.74 & 81.20 & 9.15 & 3.24 & 31.18 & 38.46 & 53.83 & 34.80 & 10.77 \\
Qwen3-Omni-30B-A3B & 21.33 & 77.16 & 15.53 & 6.54 & 45.66 & 66.67 & 18.10 & 33.15 & 9.50 \\
Step-Audio-R1.1 & 29.84 & 68.34 & 65.25 & 7.32 & 41.98 & 20.51 & 22.64 & 49.86 & 51.18 \\
\midrule
Qwen3.5-Omni-Plus & 52.77 & 47.52 & \textbf{5.13} & 19.55 & 74.88 & \underline{70.27} & \textbf{77.82} & 86.22 & 60.42 \\
StepAudio 3 Chat Preview & \underline{57.12} & \underline{44.20} & 6.85 & 31.67 & \underline{75.27} & 56.14 & 73.26 & \underline{86.29} & 64.52 \\
Doubao-Seed-2.1-Lite & 33.36 & 70.90 & 7.44 & 11.06 & 65.49 & 59.38 & 30.16 & 50.86 & 37.84 \\
Gemini 3.1 Pro Preview & 52.75 & 48.61 & 12.52 & \textbf{43.31} & 64.66 & 46.51 & 55.38 & 74.78 & 68.99 \\
\midrule
\textbf{Xing-Audio-NVVASR} & \textbf{60.36} & \textbf{37.21} & \underline{6.10} & \underline{33.95} & \textbf{80.24} & \textbf{76.32} & \underline{75.25} & \textbf{88.81} & \textbf{75.78} \\
% END GENERATED table_detail_XYZ-1800
\bottomrule
\end{tabular*}
\caption{XYZ-1800 pooled diagnostics and per-class F1.}
\label{tab:detail-xyz}
\end{table*}

\subsection{Comparing the Two WESR Reference Versions}
\label{app:original}
Table~\ref{tab:wesr-original} compares the two WESR reference versions. Qwen3.5-Omni-Plus leads in composite score under the original references, while \model{} leads after relabeling. The audit in Section~\ref{sec:wesr} found many omitted breaths; correcting these labels provides a more complete basis for assessing breathing detection.

\begin{table*}[tp]
\centering\small\setlength{\tabcolsep}{4.5pt}\renewcommand{\arraystretch}{1.08}
\begin{tabular*}{\textwidth}{@{\extracolsep{\fill}}lrrrrrrr@{}}
\toprule
& \multicolumn{5}{c}{Original WESR-400 references} & \multicolumn{2}{c}{Relabeled references}\\\cmidrule(lr){2-6}\cmidrule(l){7-8}
Model & Score & Macro-F1 & Breath P & Breath R & Breath F1 & Score & Breath F1\\
\midrule
% BEGIN GENERATED table_wesr_original
Multilingual-NVASR & 62.67 & 44.38 & \textbf{62.04} & 49.42 & \textbf{55.02} & 51.68 & 36.02 \\
MOSS-Audio-8B-Instruct & 43.52 & 38.05 & 52.78 & 11.05 & 18.27 & 34.90 & 8.79 \\
Qwen3-Omni-30B-A3B & 25.01 & 18.56 & 8.46 & 15.79 & 11.02 & 22.94 & 10.14 \\
Step-Audio-R1.1 & 28.19 & 23.76 & 21.28 & 11.63 & 15.04 & 22.05 & 7.62 \\
\midrule
Qwen3.5-Omni-Plus & \textbf{68.47} & \textbf{65.11} & \underline{61.06} & 40.12 & \underline{48.42} & 55.50 & 25.77 \\
StepAudio 3 Chat Preview & \underline{67.01} & 61.29 & 41.00 & 56.98 & 47.69 & 60.38 & 40.91 \\
Doubao-Seed-2.1-Lite & 57.69 & 53.14 & 38.46 & 41.18 & 39.77 & 53.11 & 36.07 \\
Gemini 3.1 Pro Preview & 52.59 & 56.04 & 20.77 & \textbf{75.00} & 32.53 & \underline{61.36} & \underline{51.40} \\
\midrule
\textbf{Xing-Audio-NVVASR} & 65.10 & \underline{63.62} & 35.36 & \underline{74.42} & 47.94 & \textbf{66.40} & \textbf{57.34} \\
% END GENERATED table_wesr_original
\bottomrule
\end{tabular*}
\caption{WESR results under original and relabeled references.}
\label{tab:wesr-original}
\end{table*}

\subsection{WESR Relabeling and the Evaluation of Human Supervision}
\label{app:human-comparison}
Tables~\ref{tab:human-original} and~\ref{tab:human-relabeled} revisit the decline that prompted our WESR audit. \textsc{Base}+Human-24k adds 24,000 human-annotated training utterances with other conditions fixed and corresponds to \model{}.

Adding human supervision lowers F1 in all six classes under the original references, but improves four under the relabeled references. Breathing gains 20.84 percentage points, and the composite score gains 5.97 points; crying and sighing remain lower. Together with the listening-based audit, this comparison shows how test-label omissions can obscure improvements from human supervision, motivating the more complete WESR-400-Relabeled benchmark.

\begin{table*}[tp]
\centering\small\setlength{\tabcolsep}{4pt}
\begin{tabular}{lrrrrrrrr}
\toprule
Variant & Score & Macro-F1 & Breath & Laugh & Cry & Cough & Throat & Sigh\\
\midrule
\textsc{Base} & \textbf{72.69} & \textbf{70.22} & \textbf{57.23} & \textbf{82.04} & \textbf{63.64} & \textbf{76.29} & \textbf{75.47} & \textbf{66.67} \\
\textsc{Base}+Human-24k & \underline{65.10} & \underline{63.62} & \underline{47.94} & \underline{81.17} & \underline{48.28} & \underline{70.00} & \underline{74.58} & \underline{59.74} \\
\bottomrule
\end{tabular}
\caption{Effect of additional human supervision under the original WESR references.}
\label{tab:human-original}
\end{table*}

\begin{table*}[tp]
\centering\small\setlength{\tabcolsep}{4pt}
\begin{tabular}{lrrrrrrrr}
\toprule
Variant & Score & Macro-F1 & Breath & Laugh & Cry & Cough & Throat & Sigh\\
\midrule
\textsc{Base} & \underline{60.43} & \underline{61.69} & \underline{36.50} & \underline{78.10} & \textbf{54.90} & \underline{72.29} & \underline{75.00} & \textbf{53.33} \\
\textsc{Base}+Human-24k & \textbf{66.40} & \textbf{63.91} & \textbf{57.34} & \textbf{78.18} & \underline{51.06} & \textbf{76.36} & \textbf{77.42} & \underline{43.08} \\
\bottomrule
\end{tabular}
\caption{Effect of additional human supervision under the relabeled WESR references.}
\label{tab:human-relabeled}
\end{table*}

\end{document}